\documentclass[runningheads]{llncs}
\usepackage{cite}
\usepackage{amsmath,amssymb,amsfonts}
\usepackage{graphicx}
\usepackage{textcomp}
\usepackage{xcolor}
\usepackage{multirow}
\usepackage{booktabs}
\usepackage{bm}
\usepackage{makecell}
\usepackage[linesnumbered,ruled,vlined]{algorithm2e}
\usepackage[normalem]{ulem}

\def\BibTeX{{\rm B\kern-.05em{\sc i\kern-.025em b}\kern-.08em
    T\kern-.1667em\lower.7ex\hbox{E}\kern-.125emX}}

\newcommand{\R}{{\mathbb R}}

\newcommand{\be}{\begin{equation}}
\newcommand{\ee}{\end{equation}}
\newcommand{\ba}{\begin{array}}
\newcommand{\ea}{\end{array}}
\newcommand{\baa}{\left[\begin{array}}
\newcommand{\eaa}{\end{array}\right]}

\newcommand{\beqa}{\begin{eqnarray}}
\newcommand{\eeqa}{\end{eqnarray}}
\newcommand{\bt}{\begin{tabular}}
\newcommand{\et}{\end{tabular}}

\newcommand{\bi}{\begin{itemize}}
\newcommand{\ei}{\end{itemize}}
\newcommand{\bc}{\begin{center}}
\newcommand{\ec}{\end{center}}

\begin{document}

\title{Atom dimension adaptation for infinite set dictionary learning
}

\author{Andra Băltoiu\inst{1}\orcidID{0000-0003-3600-0531} 
\and Denis C. Ilie-Ablachim\inst{1}\orcidID{0009-0005-6984-2171} 
\and Bogdan Dumitrescu\inst{1}\orcidID{0000-0003-4555-1714}}

\institute{Faculty of Automatic Control and Computers, National University of Science and Technology Politehnica Bucharest \email{\{andra.baltoiu, denis.ilie\_ablachim, bogdan.dumitrescu\}@upb.ro}}

\maketitle

\begin{abstract}
Recent work on dictionary learning with set-atoms has shown benefits in anomaly detection.
Instead of viewing an atom as a single vector, these methods allow building sparse representations with atoms taken from a set around a central vector; the set can be a cone or may have a probability distribution associated to it.
We propose a method for adaptively adjusting the size of set-atoms in Gaussian and cone dictionary learning.
The purpose of the algorithm is to match the atom sizes with their contribution in representing the signals.
The proposed algorithm not only decreases the representation error, but also improves anomaly detection, for a class of anomalies called `dependency'.
We obtain better detection performance than state-of-the-art methods.
\end{abstract}


\section{Introduction}

Dictionary learning (DL) is a class of signal processing methods that approach signal approximation by constructing a sparse representation.
A signal $\bm{y} \in \R^m$, belonging to a given set of signals, is approximated by a linear combination of the columns of an overcomplete basis, the dictionary $\bm{D} \in \R^{m \times n}$, namely by minimizing $\| \bm{y} - \bm{Dx}\|$ with respect to both $\bm{D}$ (over the whole set of signals) and the representation $\bm{x} \in \R^n$.
The representation usually contains only $s \ll m$ non-zero elements.

DL algorithms are well-established in the fields of compressed sensing \cite{DuSap09} and image processing \cite{ElAh06} and results in these domains promoted further developments including federated learning \cite{LiShi23} and found numerous applications such as texture classification \cite{RaSp10}, time-series imputation \cite{DumGiu22}.

Whether they are errors, faults, novelties or frauds, anomalies represent instances or events that differ significantly from the expected normal observations in a particular application. 
Most suited for the task of anomaly detection, namely the identification of these unusual samples in data, are methods that work in an unsupervised manner, since the outliers are uncommon in kind as well as in frequency. 

Recently, DL has been used to solve the anomaly detection problem for different types of signals, such as video \cite{YuFe18}, electrocardiogram signals \cite{AdEl15}, satellite data \cite{PBET20} or network traffic \cite{Kie20}.
A related approach is that of subspace learning, in which a set of dictionaries is learned for representing different signal features. 
The method is suited for anomaly detection \cite{TuLiu24}, since in many applications the normal signals share common features.

Roughly all DL solutions to the anomaly detection problem use the representation error as indicator (score) of abnormality; they assume that the number of normal samples and their similarity drive the learning process towards a better representation of normalcy.
While this assumption does not hold for all types of anomalies \cite{BID24_cone}, the approach remains suited for a significant amount of real-world applications, as the above cited works show.

{\em Contribution.} The present proposal builds on \cite{BID24_cone,IBD24_gauss}, which have shown that using atoms (i.e. the columns of the dictionary) that are infinite sets instead of vectors leads to better representations, essentially due to them covering a larger space. 
There, we have argued that framing DL with set-atoms creates an asymmetric problem, since efficiency imposes that only one representative of each set is retained and no reconstruction is further possible.
In short, this asymmetry implies that given a signal, the representation can be computed if the dictionary containing the central atoms is known;
however, given the representation and the dictionary, the original signal cannot be recovered.
The benefits of a more exact approximation outweigh this apparent limitation and make the approach suited, among others, for anomaly detection, as results in the above works show.

Since normal signals are more similar to each other than they are to anomalies, a conveniently placed infinite set-atom yields small representation errors for the normal signals in its vicinity.
The dimensions of the sets are therefore critical to the ability of discerning between normal samples and anomalies.

We propose a method for adapting the size of the set-atoms that is informed of the contribution each atom has in representing the samples.
Since we use DL in an unsupervised setup, we rely on atom use as an indicator of the amount of space each atom should cover in order for learning to produce discriminative results for normal and abnormal signals.

\section{Dictionary learning with set-atoms}

When working with set-atoms, the (normalized) columns $\bm{d}_j$, $j=1:n$, of the dictionary $\bm{D} \in \R^{m \times n}$ are called central atoms.
The actual atoms used to represent a signal $\bm{y} \in \R^m$ are denoted $\bm{a}_j$; they are related to the central atoms in two ways.
In the first, presented in \cite{BID24_cone}, the actual atom $\bm{a}_j$ belongs to the cone (or, more precisely, spherical cap)
\[
{\cal C}(\bm{d}_j, \rho_j) = \{ \bm{a}_j \in \R^m\ | \ \|\bm{a}_j - \bm{d}_j\| \le \rho_j, \ \| \bm{a}_j \|=1 \},
\]
where $\rho_j$ can be interpreted as the radius of the cone.

The sparse representation problem is formulated as
\be
\ba{ccl}
\underset{\bm{x} \in \R^n}{\text{min}}
& & \|\bm{y}-\sum_{j=1}^n \bm{a}_j x_j \|_2 \\
\text{s.t.}
& & \|\bm{x}\|_{0} \le s \\
& & \bm{a}_j \in {\cal C}(\bm{d}_j, \rho_j), \ j=1:n
\ea
\label{sparse_repr_cone}
\ee
where $s$ is the given sparsity level.
Problem \eqref{sparse_repr_cone} can be solved with an algorithm presented in \cite{cone_omp} and named Cone-OMP.

The dictionary is therefore composed of the central atoms, each having an associated radius.
As previously noted, the actual atoms used in the representations need not be stored.

In the second representation with set-atoms, the columns $\bm{d}_j$ are seen as the centers of normal distributions ${\cal N}(\bm{d}_j, \rho_j \bm{I})$, where $\rho_j$ are standard deviations and the covariance matrix is $\rho_j^2 \bm{I}$.
The representation problem is based on a trade-off between the probability of the actual atoms and the representation error.
The objective function is
\be
\sum_{j=1}^n \frac{1}{\rho_j^2} \|\bm{a}_j - \bm{d}_j \|^2 + 
\lambda \|\bm{y}-\sum_{j=1}^n x_j \bm{a}_j \|^2 + \gamma \|\bm{x}\|_1
\label{sparse_repr_gauss}
\ee
where $\lambda,\gamma>0$ are constants; the third term of the objective has the usual role of inducing sparsity.
The constraints $\|\bm{a}_j\|=1$, $j=1:n$, have to be imposed.
Problem \eqref{sparse_repr_gauss} can be efficiently solved with the algorithm Gauss-L1 presented in \cite{IBD24_gauss}.

We stress that problems \eqref{sparse_repr_cone} and \eqref{sparse_repr_gauss} are related to the central atoms in two distinct ways, leading to two distinct DL formulations.

In the dictionary learning problem, $N$ training signals arranged in the matrix $\bm{Y} \in \R^{m \times N}$ are given and the dictionary $\bm{D}$ of central atoms is desired.
The existent DL algorithms have the usual iterative form that alternates representation and dictionary update steps.

Cone-DL \cite{BID24_cone} uses Cone-OMP for representation and an atom update rule that implicitly combines averaging and soft thresholding.
DL-Gauss-L1 \cite{IBD24_gauss} uses Gauss-L1 for representation and simple averaging of the actual atoms corresponding to the same central atom for atom update.
Both update rules work on the fly: the actual atoms are used immediately after their computation and can be then discarded.

The above two DL algorithms assume that the cone radii and standard deviations $\rho_j$ are rigidly assigned to the central atoms.
(For brevity, we will use the word `radii' also for the standard deviations.)
Our purpose here is to allow more flexibility.
More precisely, although we still assume that the set of radii is given, we allow the one-to-one mapping radii-to-atoms to take any form in the aim to minimize the DL objective.
Recall that the radius controls the volume that an atom can cover when representing signals. 
Our approach is to let the learning process inform the need for larger or smaller radii for each atom.

\section{Atom size adaptation}

\subsection{Gaussian atoms}

We explain the problem and the solution first for Gaussian atoms.
We are given the training matrix $\bm{Y} \in \R^{m \times N}$ and the set of radii $\rho_j$, $j=1:n$.
An initial dictionary $\bm{D}_0 \in \R^{m \times n}$ is also available.
The goal is to find a dictionary $\bm{D} \in \R^{m \times n}$ and a permutation $\pi$ of $1:n$ such that
\be
\sum_{\ell=1}^N \left[ \sum_{i=j}^n \frac{1}{\rho_{\pi_j}^2} \|\bm{a}_{j\ell} - \bm{d}_{j \ell} \|^2 + 
\lambda \|\bm{y}_\ell-\sum_{j=1}^n x_{j\ell} \bm{a}_{j \ell} \|^2 + \gamma \|\bm{x}_\ell\|_1
\right]
\ee
is minimum.
So, the minimization of the trade-off between representation error, probability of actual atoms, and sparsity of coefficients must be performed by manipulating the central atoms $\bm{d}_j$ and the permutation that associates radius $\bm{\rho}_{\pi_j}$ to atom $\bm{d}_j$.
The actual atoms are also variables, but they appear in decoupled manner: atom $\bm{a}_{j\ell}$ is used only in the representation of signal $\bm{y}_\ell$.

We propose an algorithm that trains the dictionary by starting with equal radii for all atoms and slowly moving them towards their given distribution, while periodically permuting the radii such that the most used atoms get the larger radii.
The algorithm has the usual iterative form.
In each iteration, there are three main operations:
\bi
  \item Representation: for given central atoms $\bm{d}_j$ and permutation $\pi$, find actual atoms $\bm{a}_{j\ell}$ and coefficients $x_{j\ell}$ by solving problem \eqref{sparse_repr_gauss} with algorithm Gauss-L1.
  \item Central atoms update: the new atoms are computed with
  \be
    \bm{d}_j = \sum_{\ell, x_{jl}\ne 0} \bm{a}_{j\ell},
  \label{atom_update}
  \ee
  followed by normalization. So, the new central atoms are the average of the actual atoms that are effectively used in representations.
  This is the update rule of algorithm DL-Gauss-L1 \cite{IBD24_gauss}.
  \item Radii permutation update. This is the proposed addition to the algorithm and we describe it below in detail.
\ei

First of all, the update of the radii need not be made at each iteration; we make it every $\nu$-th iteration, for saving operations and letting the central atoms more time to settle in the current radii configuration.
The optimal choice of $\nu$, with respect to the performance-complexity trade-off, may be determined empirically; however we consider the detection improvements to be marginal, particularly in the case where the total number of iterations of the DL algorithm, $n_{it}$, is sufficiently large.
We denote
$\bar{\bm{\rho}} \in \R^n$ the given vector of radii.
We start the DL algorithm with all radii being equal to the average
\be
\mu = \frac{1}{n} \sum_{j=1}^n \bar{\rho}_j.
\label{av_rad}
\ee
This distribution is gradually changed towards $\bar{\bm{\rho}}$, in steps of size
\be
\bm{\delta} = (\bar{\bm{\rho}} - \mu \cdot \bm{1}) / \lfloor n_{\mathrm{it}} / \nu \rfloor,
\label{step_size}
\ee
where $\bm{1}$ is a vector whose elements are all equal to $1$.
At iteration $k$, which is a multiple of $\nu$, the radii vector has the form
\be
\bm{\rho} = \mu \cdot \bm{1} + (k/\nu) \cdot \bm{\delta}.
\label{rad_it_k}
\ee

At each radii change, the permutation $\pi$ is updated such that the radii are associated with atoms according to their use: an atom more used in representations than another atom gets a larger radius.
We employed two measures for atoms use.
The first is the number of signals in which the atom appears in representation:
\be
u_0(\bm{a}_j) = \sum_{\ell, x_{jl}\ne 0} 1.
\label{use0}
\ee
The second is the sum of the absolute value of the coefficients of the atom in the representations
\be
u_1(\bm{a}_j) = \sum_{\ell} |x_{jl}|.
\label{use1}
\ee
Relations \eqref{use0} and \eqref{use1} represent the 0-norm and the 1-norm, respectively, of the vector of coefficients of atom $\bm{a}_j$ for signals $\bm{y}_\ell$, $\ell=1:N$.
(We have also tried the 2-norm, with worse results.)
So, at each $\nu$-th iteration, the radii \eqref{rad_it_k}, sorted decreasingly, are allocated to the atoms based on the decreasing order of
$u_0$ or $u_1$.

Algorithm \ref{alg:DLG_adapt} summarizes the above operations.
We name it DL-G-L1-adapt, acknowledging that it is an improvement of DL-Gauss-L1 \cite{IBD24_gauss}.
The number of operations is only slightly larger than that of DL-Gauss-L1.

\begin{algorithm}[t]
	\SetKwComment{Comment}{}{}
	
	\BlankLine
	\KwData{training signals $\bm{Y} \in \R^{m \times N}$ \\
		\hspace{8.5mm} initial central atoms $\bm{D} \in \R^{m \times n}$ \\
		\hspace{8.5mm} radii vector $\bar{\bm{\rho}} \in \R^n$ \\
		\hspace{8.5mm} number of iterations $n_{\mathrm{it}}$
	}
	\KwResult{trained dictionary $\bm{D} \in \R^{m \times n}$ \\
		\hspace{8.5mm} permutation $\pi$
	}
	\BlankLine
	Sort $\bar{\bm{\rho}}$ decreasingly \\
    Compute step size $\bm{\delta}$ with \eqref{av_rad}, \eqref{step_size} \\
	\For{$k = 1$ \KwTo $n_{\mathrm{it}}$}{
		\For{$\ell = 1$ \KwTo $N$}{
			Compute representation $\bm{x}_\ell$ and actual atoms $\bm{a}_{j\ell}$ for $\bm{y}_{\ell}$, solving \eqref{sparse_repr_gauss} with Gauss-L1
		}
        Update central atoms with \eqref{atom_update} and normalize \\
        \If{$\bmod(k,\nu)=0$}{
            Compute atoms use with \eqref{use0} or \eqref{use1} \\
            $\pi$ is the permutation that sorts decreasingly the atom use values \\
            Update radii: $\rho_j \leftarrow \bar{\rho}_{\pi_j}$, $j=1:n$
        }
	}
	\caption{DL-G-L1-adapt}
\label{alg:DLG_adapt}
\end{algorithm}

\subsection{Cone atoms}

Radii adaptation for cone atoms is similar with that for Gaussian atoms presented above.
The resulting algorithm, DLC-adapt (C stands for cone), has the general structure of Algorithm \ref{alg:DLG_adapt}, with some modifications described below.

Of course, in step 5 the representation algorithm is Cone-OMP \cite{cone_omp} and the representation problem is \eqref{sparse_repr_cone}.
The central atoms update in step 6 is different, see \cite{BID24_cone}.
The mechanism for radii adaptation is the same, with an important addition.

Re-assigning the radii in step 10 according to either of the two atom use methods might result in cone superposition.
This problem is irrelevant for Gaussian atoms, but has to be forbidden for cone atoms, due to ambiguity of representation.
Superposition can happen especially in the case when the signals cover only a small subspace in the $m$-dimensional representation space or when the newly assigned radius size for a given atom differs significantly from its previous value.

We follow each radii re-distribution step with a procedure for distancing the overlapping atoms away from one another. 
The method is similar to how \cite{MBP12ink} approaches atom decorrelation.
We search for all pairs $(\bm{d}_1, \bm{d}_2)$ of overlapping atoms, namely for which \cite{BID24_cone}
\be
\| \bm{d}_1 - \bm{d_2} \| < \rho_1 \sqrt{1 - \frac{\rho_2^2}{4}} + \rho_2 \sqrt{1 - \frac{\rho_1^2}{4}} + \delta_0
\label{overlap_cones}
\ee
where $\delta_0$ is an imposed minimum separation threshold between the cones.
The quantity on the right represents the minimum desired distance between the two central atoms, denoted $\delta_{min}$. Also, let $\Delta = \| \bm{d}_1 - \bm{d_2} \|$. 

In order to obtain two new atoms, $\hat{\bm{d}}_1$ and $\hat{\bm{d}}_2$, such that $\| \hat{\bm{d}}_1- \hat{\bm{d}}_2 \| = \delta_{min}$, we rotate $\bm{d}_1$ and $\bm{d_2}$ away from one another in the plane generated by them, with the angle 
\be
\theta = \arcsin \left (\frac{\delta_{min}}{2} \sqrt{1- \frac{\Delta^2}{4}} - \frac{\Delta}{2} \sqrt{1-\frac{\delta_{min}^2}{4}} \right).
\label{rotation}
\ee

The procedure is repeated until all pairs of overlapping atoms are separated (note that distancing a pair of atoms might cause another pair to overlap).

Changing atom directions indiscriminately could affect their representation accuracy. 
Since the radius associated with a particular atom reflects its significance in the representation (larger radii corresponding to either frequently used or large weighted atoms), a solution is to replace the symmetric rotation in \eqref{rotation} such that the angle by which each atom is rotated is inversely proportional to its radius.

\section{Numerical experiments}

We compare our algorithms\footnote{We provide a Matlab implementation of Algorithm \ref{alg:DLG_adapt} and other relevant information at \url{https://asydil.upb.ro/software/}.} 
with those from ADBench \cite{adbench} in anomaly detection, namely with 14 unsupervised methods suited for tabular data.
Note that while ADBench contains 30 algorithms, we exclude supervised and semi-supervised methods from our tests because the DL formulation that we base our work on is unsupervised in nature. 
On a different note, due to the character of most real-world anomaly detection problems, (semi-)supervised approaches are impractical, since the case where examples of all types of anomalies are available rarely applies.

In the usual DL for anomaly detection setup, the dictionary is learned on the training set (in an unsupervised manner), then representation errors on the test set are used to classify anomalies.
This follows the intuition that due to the outnumbering of anomalies by the normal signals, the dictionary will be better specialized in representing the normal class.

We test the methods on 30 datasets from ADBench, which cover a wide range of domains, signal types and dimensions. 
For the complete list of these datasets see \cite{BID24_cone}.
Note that ADBench contains 54 datasets.
We exclude the 10 computer vision (CV) and natural language processing (NLP) datasets from our tests.
ADBench does not contain raw data of these sets, rather it uses deep learning methods to extract features from the original datasets.
While the authors argue that this approach generally improves anomaly detection, we limit our tests to unprocessed data.
For brevity, we further exclude 14 datasets, without hindering on domain diversity or anomaly percentage.

ADBench allows for synthetic dataset generation with different types of anomalies.
This is achieved by stripping the real datasets of anomalies and injecting new ones such that they present certain properties.
The different types of anomalies are related to the extent in which anomalies differ from the normal samples.
We choose the dependency type for our tests, as it is both challenging and suited for the use of representation error as anomaly score.
For each dataset, 70\% of the dataset is used for training the DL algorithm and 30\% for testing, where the representations are computed using Gauss-L1 for Gaussian atoms and Cone-OMP for cone atoms. 
Note that this train/test ratio is the same as reported in ADBench \cite{adbench}.

{\em Performance indicators.}
We report the ROC AUC (Receiver Operating Characteristic
Area Under Curve) values as performance indicators, averaged on the 30 datasets.

Moreover, we compute the rank of our proposed methods among the 14 ADBench methods, again averaged on all datasets.
The rank is individually computed for each proposed method; hence, it has values between 1 (the best) and 15.

{\em Algorithms.}
For Gaussian atoms, we use the following DL algorithms.
Remind that the set of radii is given.
The first is simply AK-SVD \cite{rubinstein2008efficient}, which trains the dictionary of central atoms in standard manner; the radii are allocated randomly to atoms; the algorithm bears the name of Gauss-L1 \cite{IBD24_gauss}, which is used for representation of test signal.
DL-Gauss-L1 is initialized with the dictionary produced by AK-SVD and radii allocated randomly to atoms and optimizes the objective function \eqref{sparse_repr_gauss}.
Finally, DL-G-L1-adapt is the algorithm proposed here for DL with radii adaptation; it is initialized with the AK-SVD dictionary and it can use the 0-norm \eqref{use0} or the 1-norm \eqref{use1} for radii reallocation.

Similarly, for cone atoms, Cone-DL is the algorithm from \cite{BID24_cone}, initialized with a random dictionary and radii allocated randomly to atoms.
DLC-adapt is the algorithm proposed here.
The letter `D' added to the name of the algorithm marks the initialization with the dictionary trained by AK-SVD.

As a baseline for DL algorithms, we use also AK-SVD followed by the representation of test signals with OMP \cite{omp}; again, representation error is used as anomaly score; we name this method AK-SVD + OMP.
Of course, the algorithms from ADBench are directly used in comparisons.

\begin{table}[t]
\caption{Average ROC AUC values for Gaussian DL algorithms}
\label{tab:g_roc_auc}
\bc
\bt{l|c|ccc|ccc}
& & \multicolumn{3}{|c|}{$\rho_{\min} = 0.01$, $\rho_{\max} = 0.1$} &
\multicolumn{3}{|c}{$\rho_{\min} = 0.04$, $\rho_{\max} = 0.12$} \\
\cline{3-8}
& $n/m$ & linear & \makecell{min-max \\50-50} & \makecell{min-max \\80-20} & linear & \makecell{min-max \\50-50} & \makecell{min-max \\80-20} \\
\hline
            & 2   & $0.9511$ & $0.9506$ & $0.9505$ & $0.9511$ & $0.9503$ & $0.9509$ \\
Gauss-L1    & 2.5 & $0.9522$ & $0.9519$ & $0.9517$ & $0.9522$ & $0.9515$ & $0.9519$ \\
            & 3   & $0.9549$ & $0.9548$ & $0.9542$ & $0.9550$ & $0.9545$ & $0.9545$ \\
\hline
            & 2   & $0.9531$ & $0.9520$ & $0.9517$ & $0.9538$ & $0.9515$ & $0.9530$ \\
DL-Gauss-L1 & 2.5 & $0.9547$ & $0.9537$ & $0.9529$ & $0.9556$ & $0.9509$ & $0.9543$ \\
            & 3   & $0.9569$ & $0.9560$ & $0.9549$ & $0.9574$ & $0.9543$ & $0.9561$ \\
\hline
            & 2   & $0.9557$ & $0.9556$ & $0.9543$ & $0.9567$ & $0.9562$ & $0.9564$ \\
DLG-L1-adapt& 2.5 & $0.9566$ & $0.9569$ & $0.9548$ & $0.9579$ & $0.9579$ & $0.9570$ \\
 (0-norm)   & 3   & $0.9586$ & $0.9585$ & $0.9570$ & $\mathbf{0.9598}$ & $\mathit{0.9594}$ & $0.9588$ \\
\hline
            & 2   & $0.9565$ & $0.9559$ & $0.9555$ & $0.9568$ & $0.9566$ & $0.9574$ \\
DLG-L1-adapt& 2.5 & $0.9572$ & $0.9569$ & $0.9561$ & $0.9579$ & $0.9576$ & $0.9576$ \\
 (1-norm)   & 3   & $\mathit{0.9591}$ & $0.9583$ & $0.9584$ & $\mathit{0.9596}$ & $0.9585$ & $\mathbf{0.9598}$ \\
\et
\ec
\end{table}

\begin{table*}[t]
\caption{Average ranks for Gaussian DL algorithms}
\label{tab:g_rank}
\bc
\bt{l|c|ccc|ccc}
& & \multicolumn{3}{|c|}{$\rho_{\min} = 0.01$, $\rho_{\max} = 0.1$} &
\multicolumn{3}{|c}{$\rho_{\min} = 0.04$, $\rho_{\max} = 0.12$} \\
\cline{3-8}
& $n/m$ & linear & \makecell{min-max \\50-50} & \makecell{min-max \\80-20} & linear & \makecell{min-max\\ 50-50} & \makecell{min-max \\80-20} \\
\hline
            & 2   & $1.90$ & $1.97$ & $1.97$ & $1.90$ & $1.93$ & $1.93$ \\
Gauss-L1    & 2.5 & $1.63$ & $1.67$ & $1.70$ & $1.63$ & $1.63$ & $1.67$ \\
            & 3   & $1.53$ & $1.63$ & $1.70$ & $1.63$ & $1.63$ & $1.67$ \\
\hline
            & 2   & $1.83$ & $1.83$ & $1.90$ & $1.73$ & $1.87$ & $1.87$ \\
DL-Gauss-L1 & 2.5 & $1.57$ & $1.60$ & $1.67$ & $1.53$ & $1.63$ & $1.57$ \\
            & 3   & $1.50$ & $1.60$ & $1.63$ & $1.50$ & $1.60$ & $1.53$ \\
\hline
            & 2   & $1.60$ & $1.70$ & $1.63$ & $1.63$ & $1.60$ & $1.60$ \\
DLG-L1-adapt& 2.5 & $1.40$ & $1.43$ & $1.53$ & $1.47$ & $1.47$ & $\mathbf{1.33}$ \\
 (0-norm)   & 3   & $1.40$ & $1.43$ & $1.43$ & $1.43$ & $1.50$ & $\mathbf{1.33}$ \\
\hline
            & 2   & $1.60$ & $1.60$ & $1.60$ & $1.57$ & $1.63$ & $1.57$ \\
DLG-L1-adapt& 2.5 & $1.40$ & $1.47$ & $1.43$ & $1.43$ & $1.53$ & $1.37$ \\
 (1-norm)   & 3   & $1.43$ & $1.47$ & $1.43$ & $1.40$ & $1.43$ & $\mathbf{1.33}$ \\
\et
\ec
\end{table*}


\begin{table*}
\caption{Average representation errors for Gaussian DL algorithms}
\label{tab:g_err}
\bc
\bt{l|c|ccc}
& & 
\multicolumn{3}{|c}{$\rho_{\min} = 0.04$, $\rho_{\max} = 0.12$} \\
\cline{3-5}
& $n/m$ & linear & \makecell{min-max\\ 50-50} & \makecell{min-max \\80-20} \\
\hline
            & 2   & $0.2535$ & $0.2522$ & $0.2567$ \\
Gauss-L1    & 2.5 & $0.2424$ & $0.2419$ & $0.2451$ \\
            & 3   & $0.2357$ & $0.2358$ & $0.2382$ \\
\hline
            & 2   & $0.2482$ & $0.2478$ & $0.2527$ \\
DL-Gauss-L1 & 2.5 & $0.2379$ & $0.2381$ & $0.2415$ \\
            & 3   & $0.2316$ & $0.2318$ & $0.2349$ \\
\hline
            & 2   & $0.2450$ & $0.2438$ & $0.2488$ \\
DLG-L1-adapt& 2.5 & $0.2350$ & $0.2339$ & $0.2380$ \\
 (0-norm)   & 3   & $0.2289$ & $0.2279$ & $0.2316$ \\
\hline
            & 2   & $0.2444$ & $0.2431$ & $0.2479$ \\
DLG-L1-adapt& 2.5 & $0.2345$ & $0.2334$ & $0.2373$ \\
 (1-norm)   & 3   & $0.2282$ & $0.2272$ & $0.2309$
\et
\ec
\end{table*}

{\em Parameters.}
The minimum and maximum radii, $\rho_{\min}$ and $\rho_{\max}$ have two values, $(0.01,0.1)$ and $(0.04,0.12)$.
We experimented with three radii distributions: i) a uniform grid from $\rho_{\min}$ to $\rho_{\max}$; ii) values equal only with $\rho_{\min}$ and $\rho_{\max}$, in equal proportion; iii) idem, with a 80-20 proportion.
For the algorithms with Gaussian atoms, we take $\lambda=\gamma=1$, since these values gave the best results in \cite{IBD24_gauss}.
The other parameters of Gauss-L1 and DL-Gauss-L1, not mentioned here, are like in \cite{IBD24_gauss}.
The overcompleteness factor $n/m$ is taken from the set $\{2, 2.5, 3 \}$ for Gaussian atoms and from $\{2,3,4 \}$ for cone atoms.
The sparsity level for cone algorithms is $s=2$, since it gave the best results in \cite{BID24_cone}.
The minimum distance between cones is $\delta_0=0.01$.
For DLC-adapt, we obtain similar results when using symmetric and asymmetric angles for rotating the overlapping atoms; we only report here the symmetric case.
For all the algorithms, both adaptive and non adaptive, dictionary learning is done in $n_{it}=100$ iterations. 
In the adaptive variants, we update the radii every $\nu=10$ iterations.

{\em Results.}
We first present the averaged results over all datasets. Since they differ in characteristics and domain category and because the non adaptive variants of the infinite-set DL algorithms vary significantly (from 0.7282 to 1 ROC AUC in the case of the DL-Gauss-L1 algorithm), we then analyse the results for each dataset.
All results are averaged over 5 independent runs.

We present the results of the methods for Gaussian atoms in Tables \ref{tab:g_roc_auc} and \ref{tab:g_rank}.
The best ROC AUC and rank values are written in bold.
ROC AUC values larger than $0.9590$ are in italics.
We note that radii adaptation brings benefits in all considered cases.
There is not much difference between the radii distributions, although the linear one seems to have a slight advantage.
Also, the 1-norm \eqref{use1} used for sorting radii appears to be slightly better than the 0-norm.
The results for $(\rho_{\min},\rho_{\max}) = (0.04,0.12)$ are better.

The average representation errors (per signal element) are shown in Table \ref{tab:g_err}.
Their behavior is the expected one.
They decrease with increasing radii and overcompleteness.
The linear and 50-50 distributions, which have equal average radii, give similar errors, better than those for the 80-20 distribution, which has smaller average radius.
For these combinations of parameters, smaller error correlates well with better ROC AUC.
However, further increase of the radii no longer brings benefits in terms of ROC AUC or rank, although the error continues to decrease.

For cone algorithms, the results are given in Tables \ref{tab:c_roc_auc}--\ref{tab:c_err}.
The results are below those for Gaussian atoms, but the trends are similar.
Radii adaptation is always beneficial.
Initialization with a trained dictionary usually improves the results and 1-norm is generally better than the 0-norm.

The best results for the baseline methods are as follows.
AK-SVD + OMP has a best ROC AUC of $0.9329$ and a rank of $1.93$ \cite{BID24_cone}.
The best method from ADBench (COF \cite{tang02cof}) gives a ROC AUC of $0.9274$ (interestingly, more recent methods, including some based on neural networks, give worse results).
The results of our methods are clearly superior.

While it is beyond the purpose of the paper to investigate the best parameter configuration for the ADBench methods, we did test the influence of the parameter values on the two best performing benchmark methods, COF and LOF. 
For the rest of the methods we used the implicit values from the benchmark.
Specifically, we tested for different values for the number of neighbors in both methods ($n\_neighbors \in \{10, 15, 18, 20, 22, 25\}$) as well as different leaf sizes for LOF. 
The latter did not affect performance of LOF in any way and the default value of $n\_neighbors = 20$ achived the best results.
In the case of COF slight improvements were obtained using a smaller number of neighbors (ROC AUC of $0.9299$ for $n\_neighbors = 15$ and ROC AUC of $0.9303$ for $n\_neighbors = 10$).
While these test are not exhaustive, they suggest that even with parameter optimization for the benchmark methods, our methods are still better.

Note also that the best ROC AUC in our previous work is $0.9443$ in \cite{BID24_cone} and $0.9519$ in \cite{IBD24_gauss};
the corresponding best rank values are $1.67$ and $1.60$, respectively.
So, we have visibly improved them here.

\begin{table*}
\caption{Average ROC AUC values for cone DL algorithms}
\label{tab:c_roc_auc}
\bc
\bt{l|c|ccc|ccc}
& & \multicolumn{3}{|c|}{$\rho_{\min} = 0.01$, $\rho_{\max} = 0.1$} &
\multicolumn{3}{|c}{$\rho_{\min} = 0.04$, $\rho_{\max} = 0.12$} \\
\cline{3-8}
& $n/m$ & linear & \makecell{min-max \\50-50} & \makecell{min-max \\80-20} & linear & \makecell{min-max \\50-50} & \makecell{min-max \\80-20} \\
\hline
            & 2   & $0.9355$ & $0.9315$ & $0.9389$ & $0.9304$ & $0.9262$ & $0.9341$ \\
Cone-DL     & 3   & $0.9334$ & $0.9317$ & $0.9340$ & $0.9309$ & $0.9267$ & $0.9328$ \\
            & 4   & $0.9336$ & $0.9323$ & $0.9359$ & $0.9279$ & $0.9278$ & $0.9328$ \\
\hline
            & 2   & $0.9361$ & $0.9327$ & $0.9353$ & $0.9330$ & $0.9304$ & $0.9347$ \\
Cone-DL-D   & 3   & $0.9346$ & $0.9352$ & $0.9363$ & $0.9343$ & $0.9310$ & $0.9357$ \\
            & 4   & $0.9358$ & $0.9309$ & $0.9345$ & $0.9331$ & $0.9301$ & $0.9346$ \\
\hline
            & 2   & $0.9400$ & $0.9384$ & $0.9412$ & $0.9352$ & $0.9353$ & $0.9430$ \\
DLC-adapt   & 3   & $0.9398$ & $0.9375$ & $0.9416$ & $0.9362$ & $0.9375$ & $0.9419$ \\
 (0-norm)   & 4   & $0.9399$ & $0.9367$ & $0.9421$ & $0.9354$ & $0.9345$ & $0.9406$ \\
\hline
            & 2   & $0.9428$ & $0.9418$ & $0.9466$ & $0.9386$ & $0.9387$ & $0.9464$ \\
DLC-adapt   & 3   & $0.9440$ & $0.9418$ & $0.9474$ & $0.9432$ & $0.9403$ & $0.9447$ \\
 (1-norm)   & 4   & $0.9439$ & $0.9427$ & $0.9487$ & $0.9394$ & $0.9349$ & $0.9475$ \\
\hline
            & 2   & $0.9409$ & $0.9406$ & $0.9428$ & $0.9382$ & $0.9350$ & $0.9438$ \\
DLC-D-adapt & 3   & $0.9413$ & $0.9420$ & $0.9438$ & $0.9424$ & $0.9414$ & $0.9430$ \\
 (0-norm)   & 4   & $0.9428$ & $0.9412$ & $0.9434$ & $0.9405$ & $0.9391$ & $0.9435$ \\
\hline
            & 2   & $0.9442$ & $0.9455$ & $0.9481$ & $0.9433$ & $0.9425$ & $0.9467$ \\
DLC-D-adapt & 3   & $0.9461$ & $0.9449$ & $0.9498$ & $0.9453$ & $0.9455$ & $0.9484$ \\
 (1-norm)   & 4   & $0.9454$ & $0.9461$ & $\mathit{0.9503}$ & $0.9445$ & $0.9418$ & $\mathbf{0.9506}$ \\
\et
\ec
\end{table*}

\begin{table*}
\caption{Average rank for cone DL algorithms}
\label{tab:c_rank}
\bc
\bt{l|c|ccc|ccc}
& & \multicolumn{3}{|c|}{$\rho_{\min} = 0.01$, $\rho_{\max} = 0.1$} &
\multicolumn{3}{|c}{$\rho_{\min} = 0.04$, $\rho_{\max} = 0.12$} \\
\cline{3-8}
& $n/m$ & linear & \makecell{min-max \\50-50} & \makecell{min-max \\80-20} & linear & \makecell{min-max \\50-50} & \makecell{min-max \\80-20} \\
\hline
            & 2   & $2.00$ & $2.23$ & $1.87$ & $2.30$ & $2.67$ & $2.20$ \\
Cone-DL     & 3   & $1.97$ & $1.97$ & $2.10$ & $2.03$ & $2.10$ & $2.03$ \\
            & 4   & $2.07$ & $1.97$ & $1.83$ & $2.13$ & $2.03$ & $1.97$ \\
\hline
            & 2   & $1.97$ & $2.17$ & $1.87$ & $2.13$ & $2.23$ & $2.03$ \\
Cone-DL-D   & 3   & $2.00$ & $1.97$ & $1.93$ & $1.83$ & $2.00$ & $2.03$ \\
            & 4   & $1.90$ & $2.20$ & $1.90$ & $2.00$ & $2.13$ & $1.93$ \\
\hline
            & 2   & $1.83$ & $1.87$ & $1.80$ & $2.10$ & $2.13$ & $1.77$ \\
DLC-adapt   & 3   & $1.73$ & $1.93$ & $1.83$ & $1.87$ & $1.73$ & $1.73$ \\
 (0-norm)   & 4   & $1.83$ & $1.83$ & $1.73$ & $1.97$ & $1.97$ & $1.80$ \\
\hline
            & 2   & $1.83$ & $1.73$ & $1.60$ & $2.00$ & $1.97$ & $1.63$ \\
DLC-adapt   & 3   & $1.57$ & $1.67$ & $1.67$ & $1.70$ & $1.63$ & $1.73$ \\
 (1-norm)   & 4   & $1.63$ & $1.60$ & $1.57$ & $1.77$ & $1.97$ & $1.57$ \\
\hline
            & 2   & $1.77$ & $1.83$ & $1.70$ & $1.80$ & $2.03$ & $1.73$ \\
DLC-D-adapt & 3   & $1.70$ & $1.73$ & $1.70$ & $1.67$ & $1.87$ & $1.70$ \\
 (0-norm)   & 4   & $1.60$ & $1.67$ & $1.77$ & $1.67$ & $1.77$ & $1.63$ \\
\hline
            & 2   & $1.63$ & $1.70$ & $1.57$ & $1.63$ & $1.73$ & $1.53$ \\
DLC-D-adapt & 3   & $1.57$ & $1.70$ & $1.53$ & $1.63$ & $1.70$ & $1.57$ \\
 (1-norm)   & 4   & $1.60$ & $1.57$ & $1.50$ & $1.57$ & $1.70$ & $\mathbf{1.47}$ \\
\et
\ec
\end{table*}
\begin{table}
\caption{Average representation errors for cone DL algorithms}
\label{tab:c_err}
\bc
\bt{l|c|ccc}
& & 
\multicolumn{3}{|c}{$\rho_{\min} = 0.04$, $\rho_{\max} = 0.12$} \\
\cline{3-5}
& $n/m$ & linear & \makecell{min-max \\50-50} & \makecell{min-max \\80-20} \\
\hline
            & 2   & $0.1660$ & $0.1524$ & $0.1806$ \\
Cone-DL     & 3   & $0.1528$ & $0.1480$ & $0.1657$ \\
            & 4   & $0.1426$ & $0.1376$ & $0.1539$ \\
\hline
            & 2   & $0.1690$ & $0.1637$ & $0.1863$ \\
Cone-DL-D   & 3   & $0.1520$ & $0.1477$ & $0.1679$ \\
            & 4   & $0.1423$ & $0.1372$ & $0.1570$ \\
\hline
            & 2   & $0.1648$ & $0.1615$ & $0.1770$ \\
DLC-adapt   & 3   & $0.1460$ & $0.1411$ & $0.1563$ \\
 (0-norm)   & 4   & $0.1346$ & $0.1300$ & $0.1474$ \\
\hline
            & 2   & $0.1634$ & $0.1579$ & $0.1728$ \\
DLC-adapt   & 3   & $0.1452$ & $0.1409$ & $0.1535$ \\
 (1-norm)   & 4   & $0.1331$ & $0.1289$ & $0.1433$ \\
\hline
            & 2   & $0.1587$ & $0.1563$ & $0.1728$ \\
DLC-D-adapt & 3   & $0.1418$ & $0.1383$ & $0.1535$ \\
 (0-norm)   & 4   & $0.1311$ & $0.1267$ & $0.1433$ \\
\hline
            & 2   & $0.1569$ & $0.1535$ & $0.1711$ \\
DLC-D-adapt & 3   & $0.1405$ & $0.1363$ & $0.1523$ \\
 (1-norm)   & 4   & $0.1303$ & $0.1254$ & $0.1412$ \\
\et
\ec
\end{table}

\begin{figure}[t]
\centering
\includegraphics[width=1\textwidth]{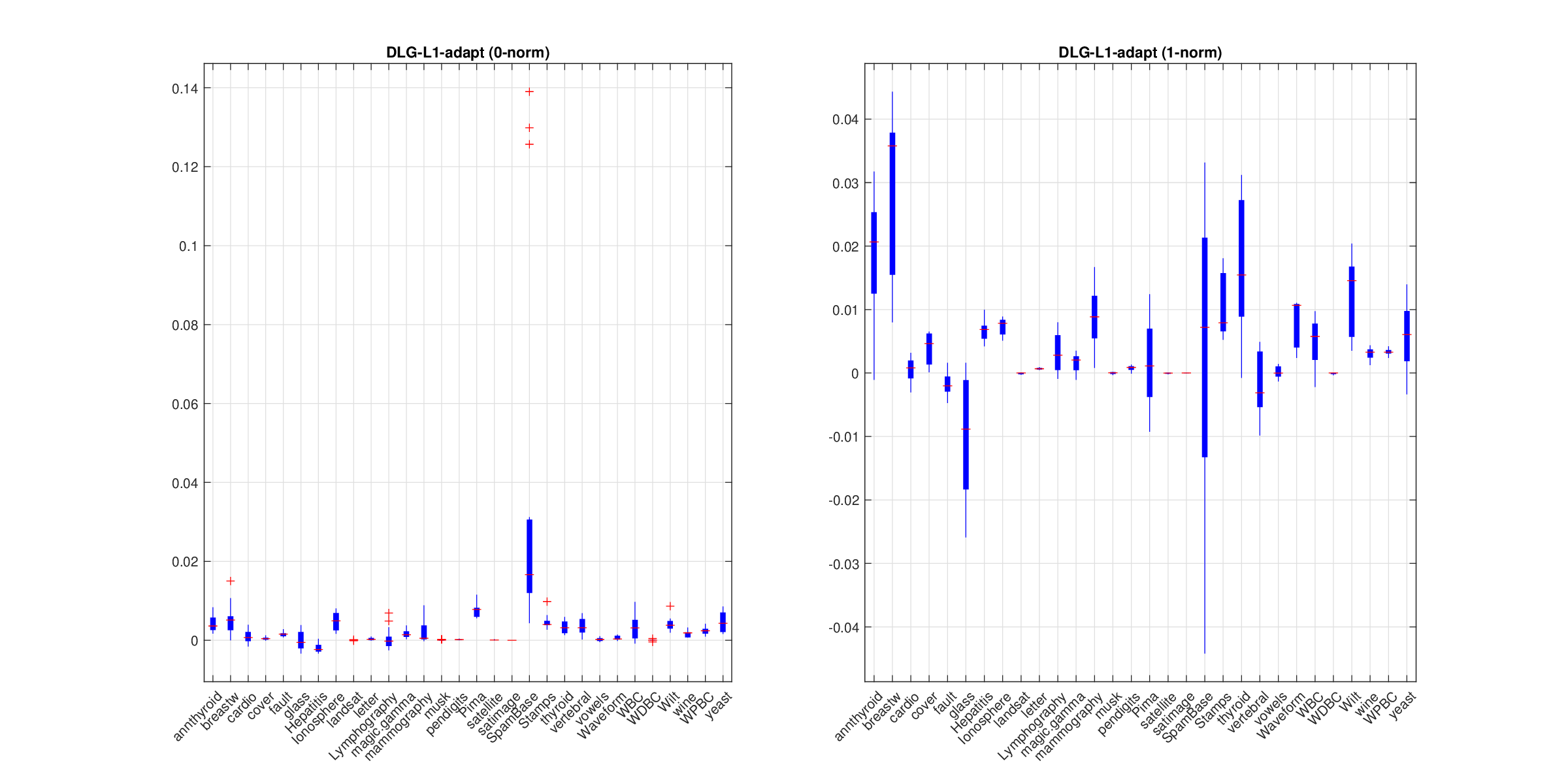} 
\caption{Box plots of improvements versus the non adaptive variant of the DL-Gauss-L1 method. For each dataset, all the 18 parameter configurations are considered ($n/m$, radii limits and distributions).}
\label{fig:difs_gauss}
\end{figure}

\begin{figure}[t]
\includegraphics[width=1.1\textwidth]{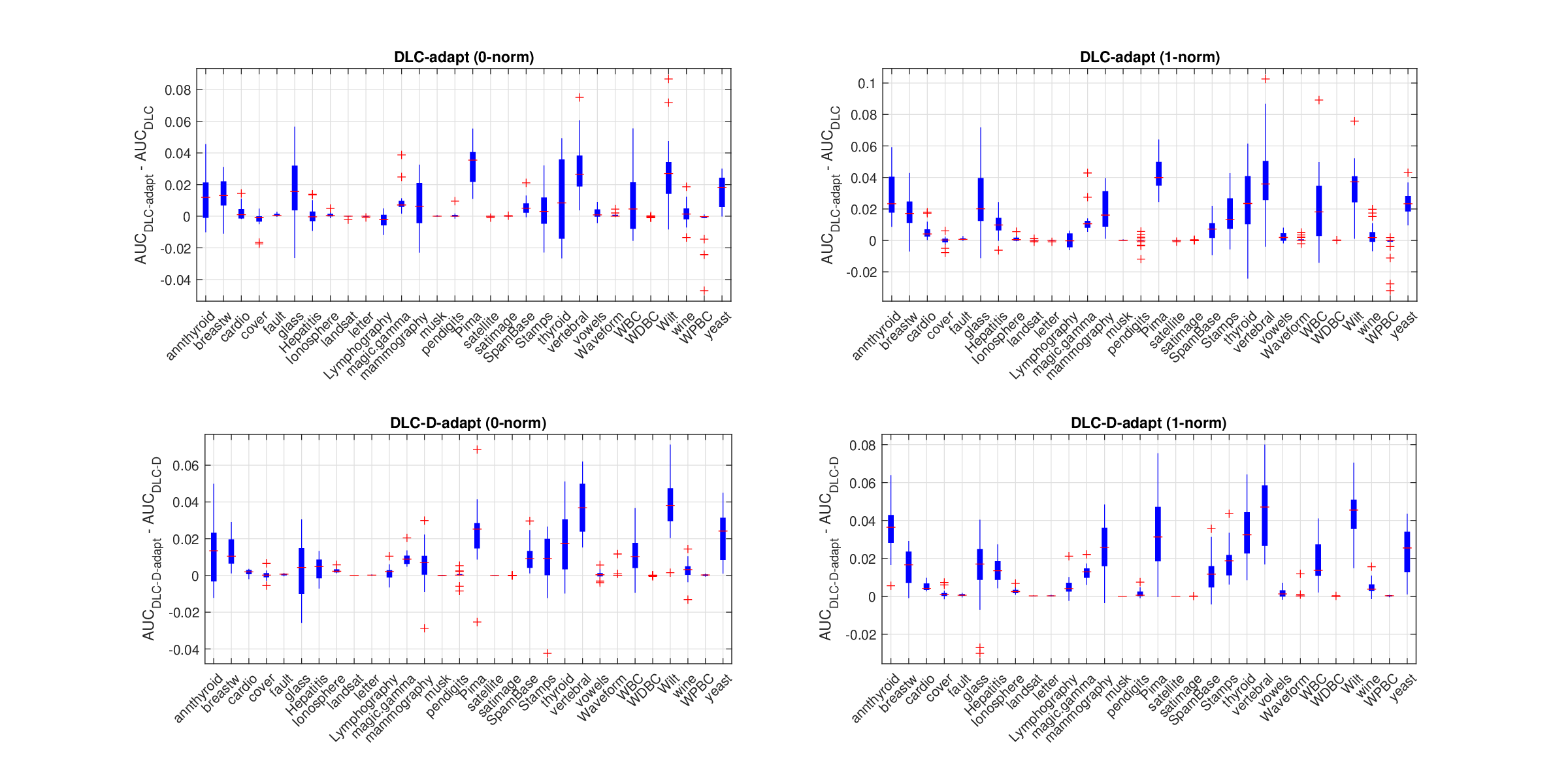} 
\caption{Box plots of improvements versus the non adaptive variants of the Cone-DL methods. For each dataset, all the 18 parameter configurations are considered ($n/m$, radii limits and distributions).}
\label{fig:difs_cone}
\end{figure}

\begin{figure}[t]
\centering
\includegraphics[width=1\textwidth]{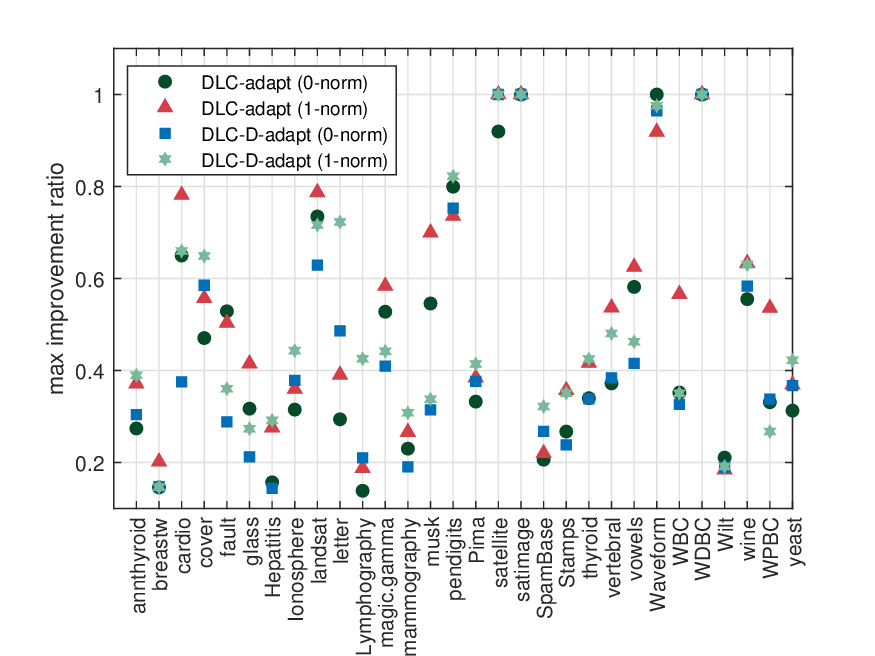} 
\caption{Maximum ratio between achieved improvement with adaptive cone DL methods and maximum possible improvement.}
\label{fig:ratio_cone}
\end{figure}

Figures \ref{fig:difs_gauss} and \ref{fig:difs_cone} present the improvements (in terms of difference in ROC AUC scores) obtained by applying the radii adaptation procedures on the cone DL and Gaussian DL algorithms, respectively, for each of the 30 datasets.
The figures present the statistics over the 18 parameter configurations (three $n/m$ values, two
radii intervals and three radii distributions). 

In the case of the Gaussian DL methods, although the 1-norm variant is more sensitive to parameter values and dataset characteristics, it also achieves better improvements than the 0-norm solution.

The cone DL solutions follow similar trends regardless of the type of norm or the usage of the cone dictionary update algorithm; while 1-norm variants are generally better, it is the dataset characteristics that largely determine the performance of the adaptive cone DL algorithms.

Since the infinite-set DL variants already have good performance on several datasets, in order to evaluate the improvement obtained by applying the adaptive variants, we analyze the increase in ROC AUC with respect to the maximum possible improvement.
We define the maximum possible improvement as the margin between the corresponding non adaptive variant and 1.
More precisely, denoting $\alpha_1$ the ROC AUC obtained by an adaptive algorithm and $\alpha_0$ the ROC AUC of the corresponding non adaptive algorithm, the proposed improvement ratio is
\be
\frac{\alpha_1 - \alpha_0}{1 - \alpha_0}.
\label{improvement_ratio}
\ee
Figure \ref{fig:ratio_cone} shows the best ratio of improvement obtained over the 18 configurations for each dataset for the cone DL algorithms.

Note that in four of the 30 datasets, the non adaptive solutions already achieve a ROC AUC of 1.
In this case, the improvement ratio is set to 1, if $\alpha_1=1$ (which always happened).
In half of the remaining 26 sets, at least $50\%$ of the improvement margin is covered by at least one of the adaptive methods.
For the datasets where this does not happen, the improvement is still significant: the mean ROC AUC improvement on these sets is $0.0487$ and the maximum is $0.0867$.


\section{Conclusions}
We propose a method for adapting the size of set-atoms for dictionary learning, which can be used both when the set is represented as a Gaussian distribution centered around a regular dictionary atom or by considering a cone atom instead of a vector.
In the context of anomaly detection, the aim of the algorithm is to adjust the standard deviations of the Gaussian distributions (or, respectively, the cone radii) such that the dictionary approximates the normal signals better than the anomalies. 
Results show that the method indeed widens the gap between the representation error for anomalies and normal signals, resulting in better ROC AUC score and an improvement of $17\%$ in the rank among ADBench algorithms, compared to previous variants of the infinite set DL methods.

\begin{credits}
\subsubsection{\ackname} This work was supported by a grant of the Ministry of Research, Innovation and Digitization, CNCS - UEFISCDI, project number PN-III-P4-PCE-2021-0154, within PNCDI III.

\subsubsection{\discintname}
The authors have no competing interests to declare that are
relevant to the content of this article.
\end{credits}

\bibliographystyle{unsrt}
\bibliography{bib}
\end{document}